\pdfoutput=1
\documentclass[11pt]{article}

\usepackage[final]{acl}

\usepackage{amsmath}
\usepackage{times}
\usepackage{booktabs}
\usepackage{graphicx}
\usepackage{threeparttable}
\usepackage{latexsym}
\usepackage[T1]{fontenc}
\usepackage[utf8]{inputenc}
\usepackage{microtype}
\usepackage{inconsolata}
\usepackage{graphicx}

\usepackage{graphicx}
\usepackage{textcomp}
\usepackage{color,soul}
\usepackage{bm}
\usepackage{xcolor}
\usepackage{colortbl}

\definecolor{ForestGreen}{RGB}{0,77,36}

\colorlet{tableheadcolor}{gray!25} 
\colorlet{tablerowcolor}{gray!15} 
\colorlet{tablerowcolor2}{gray!12} 
\colorlet{tablerowcolor3}{gray!25} 
\colorlet{tablerowcolor4}{gray!50} 

\newcommand{\rowcollight}{\rowcolor{tablerowcolor2}} %

\title{SLM-Mod: Small Language Models Surpass LLMs at Content Moderation}

\author{
    Xianyang Zhan\textsuperscript{$\ast$}, 
    Agam Goyal\textsuperscript{$\ast$}, 
    \textbf{Yilun Chen}, 
    \textbf{Eshwar Chandrasekharan}\textsuperscript{\textdaggerdbl},
    \textbf{Koustuv Saha}\textsuperscript{\textdaggerdbl} \\
    Siebel School of Computing and Data Science\\
  University of Illinois Urbana-Champaign\\
  \texttt{\{zhan39, agamg2, yilunc3, eshwar, ksaha2\}@illinois.edu}
}

\begin{document}
\maketitle
\def\thefootnote{$\ast$}\footnotetext{Both authors contributed equally.}\def\thefootnote{\arabic{footnote}}
\def\thefootnote{\textdaggerdbl}\footnotetext{Both authors are advisors of this work.}\def\thefootnote{\arabic{footnote}}

\begin{abstract}
Large language models (LLMs) have shown promise in many natural language understanding tasks, including content moderation. However, these models can be expensive to query in real-time and do not allow for a community-specific approach to content moderation. To address these challenges, we explore the use of open-source small language models (SLMs) for community-specific content moderation tasks. We fine-tune and evaluate SLMs (less than 15B parameters) by comparing their performance against much larger open- and closed-sourced models in both a zero-shot and few-shot setting. Using 150K comments from 15 popular Reddit communities, we find that SLMs outperform zero-shot LLMs at content moderation---11.5\% higher accuracy and 25.7\% higher recall on average across all communities. Moreover, few-shot in-context learning leads to only a marginal increase in the performance of LLMs, still lacking compared to SLMs. We further show the promise of cross-community content moderation, which has implications for new communities and the development of cross-platform moderation techniques. Finally, we outline directions for future work on language model based content moderation.\footnote{Code: \href{https://github.com/AGoyal0512/SLM-Mod}{https://github.com/AGoyal0512/SLM-Mod}.}
\end{abstract}

\section{Introduction}

Content moderation has become a growing area of interest for the NLP community~\cite{jurgens_just_2019} due to the rapid growth and use of social media.
The primary challenge in content moderation lies in detecting undesirable, norm-violating behavior amidst vast amounts of content posted by users.
In order to deal with large volumes of content, most platforms rely on automated tools to either directly remove norm-violating content or triage undesirable content for manual review by human moderators~\cite{chandrasekharan_crossmod_2019}.

\begin{figure}[t] 
    \centering
    \includegraphics[width=0.65\columnwidth]{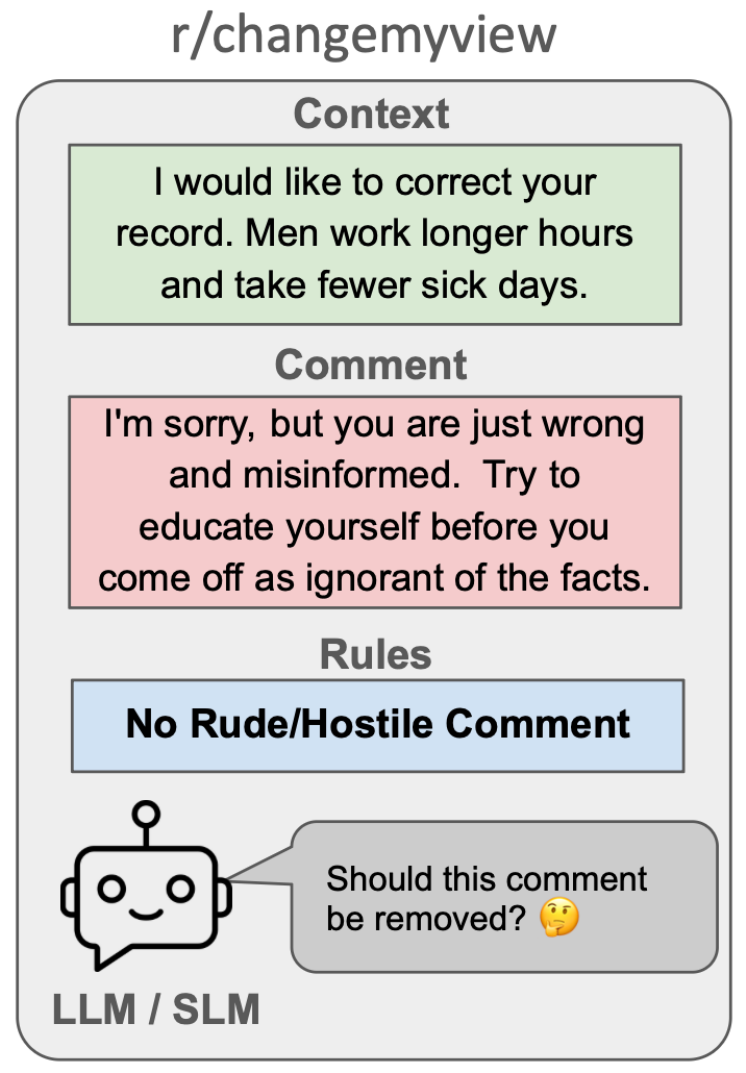} 
    \caption{\textbf{Online Moderation with Language Models.} Given a comment from a subreddit \textit{r/changemyview}, the preceding context, and community rules, we compare LLMs and SLMs on moderation performance.}
    \label{fig:teaser}
\end{figure}

 Although moderation involves dealing with a wide range of undesirable behaviors, current computational approaches tend to adopt narrow definitions of abuse (e.g., toxicity, hate speech)~\cite{jurgens_just_2019}. However, content moderation is a nuanced and contextual task that varies across platforms, communities, and over time---e.g., a post deemed desirable in one community may be undesirable in another~\cite{chandrasekharan2018internet}. 
Moderation requires an understanding of community norms, and depend on the manual labor and judgment of human moderators, who are often overworked and uncompensated~\cite{li2022measuring}. 
Despite these variations between communities, current approaches do not effectively incorporate community norms when detecting undesirable content.

Due to the natural language understanding capabilities of large language models (LLMs)~\cite{radford2019language, brown2020language}, recent work has explored the use of LLMs for content moderation~\cite{kumar2024watch, kolla2024llm}. 
Although LLMs can be used off-the-shelf and generate explanations for moderation decisions, moderators would still require the ability to dynamically adapt these models to suit the preferences and norms of their community.
In other words, there is a need for specialized models that provide moderators with more control and configurability via fine-tuning.

Current methods for fine-tuning LLMs are constrained by the substantial computational resources and costs required, making it nearly impossible for moderators to tailor LLMs for community-specific content moderation.
Given these limitations, a viable alternative could be the use of small language models (SLMs), which provide a lightweight and cost-effective option compared to LLMs. Recent research has shown that despite their smaller size, SLMs can be fine-tuned to achieve performance comparable to LLMs on various natural language understanding tasks~\cite{schick2021itsjustsizematters}. 

This work explores the potential for using SLMs in content moderation, evaluating their performance relative to LLMs in terms of accuracy, recall, and precision across multiple online communities on Reddit. 
In particular, we focus on whether fine-tuned SLMs can offer a more resource-efficient approach without compromising on moderation quality. We aim to determine if SLMs can strike a balance between cost-effectiveness and moderation accuracy, ultimately providing a viable alternative to the computationally expensive LLMs for large-scale, community-specific moderation.

\paragraph{Findings:} Our findings reveal that fine-tuned SLMs outperform zero-shot LLMs at in-domain content moderation tasks having both higher accuracy and recall, with slightly lower precision. We present a case study of \textit{r/changemyview} and conduct an error analysis on false positives and false negatives to identify trade-offs when using SLMs over LLMs. We show that even under a few-shot setting, LLMs lack performance compared to SLMs. Moreover, SLMs have a higher AUC score compared to LLMs on more realistic imbalanced datasets. Next, we highlight the potential of SLMs for cross-domain content moderation tasks and investigate possible reasons for their high performance. Finally we discuss the implications of our findings for content moderation and highlight future directions for improving cross-community approaches to moderation as well as dynamically adapting community-specific models over time.

\vspace{-8.1pt}
\section{Related Work}

\paragraph{Automated Approaches to Content Moderation:}
Due to the limited scalability of human moderation,
automated approaches have been increasingly adopted. Automated approaches typically use natural language processing and machine learning techniques to flag potentially harmful content for human review~\cite{chandrasekharan_crossmod_2019}. For example, n-gram models and sentiment analysis techniques have been used as classic automated approaches to classify content as toxic or non-toxic~\cite{davidson2017automatedhatespeechdetection, Vigna2017HateMH, warner-hirschberg-2012-detecting}. These methods rely on identifying patterns of word usage and sentiment to make judgments about the harmfulness of content, providing a foundational approach for content moderation. More advanced methods, including deep learning models 
to automatically learn multi-layers of abstract features from raw data have been explored for detecting offensive content~\cite{Nobata2016AbusiveLD,10.1145/3041021.3054223,zhang2018hatespeechdetectionsolved}. Recently,~\citet{jha2024memeguard} proposed an LLM and VLM-based framework for online content moderation via meme interventions, and~\citet{maity2023genex} developed a generative framework for explainable cyber-bullying detection.

\vspace{-6pt}
\paragraph{LLM-Assisted Content Moderation:}
The rise of large language models (LLMs) and their successors has transformed content moderation by enhancing the ability to detect harmful content with greater contextual awareness. These models excel at processing longer texts and capturing nuanced meanings, making them highly effective for identifying hate speech, misinformation, and abusive language on social platforms~\cite{kolla2024llm}. 
However, despite their advancements, LLMs are computationally expensive to run, making them less feasible for real-time moderation at scale~\cite{kumar2024watch,10.1145/3491102.3501999}. In light of these limitations, our study investigates whether SLMs, which are more lightweight and cost-effective ~\cite{10590016}, can be fine-tuned to handle moderation tasks with comparable accuracy, potentially offering a balance between cost and effectiveness in online content moderation.

\section{Experimental Setup}

\subsection{Data Curation}

We curate our data from the publicly available dataset of Reddit comment removals between May $10^{\text{th}}$, $2016$  and February $4^{\text{th}}$, $2017$ by~\citet{chandrasekharan2018internet} by sampling 10K comments (5K moderated and 5K unmoderated) for 15 popular subreddits from Reddit's landing page. 
For each subreddit, we split its data into 80/20 train/test sets. 
\autoref{app:subreddits_stats} includes the complete list, description, and subscriber statistics of the subreddits.

\subsection{Models and Configuration}

For our study, we evaluate the $4-$bit quantized versions of three small language models (SLMs): Llama-3.1-8b~\cite{dubey2024llama}, Gemma-2-9b~\cite{team2024gemma}, and Mistral-nemo-instruct~\cite{mistralMistralNeMo}, and three large language models (LLMs): Cohere's Command R+~\cite{cohere_for_ai_2024}, OpenAI's GPT-4o; and GPT-4o-mini~\cite{hellogpt4o}. For the LLMs, a \texttt{temperature} of $0$ was used to ensure consistency across runs, and a \texttt{top\_p} of $0.75$ was used.

We fine-tune 15 subreddit-specific SLMs using rank 16 Low-Rank Adaptation (LoRA)~\cite{hu2021lora} for 1 epoch on a balanced sample of total 8,000 moderated and unmoderated comments.

\begin{figure*}[t] 
    \centering
    \includegraphics[width=1\textwidth]{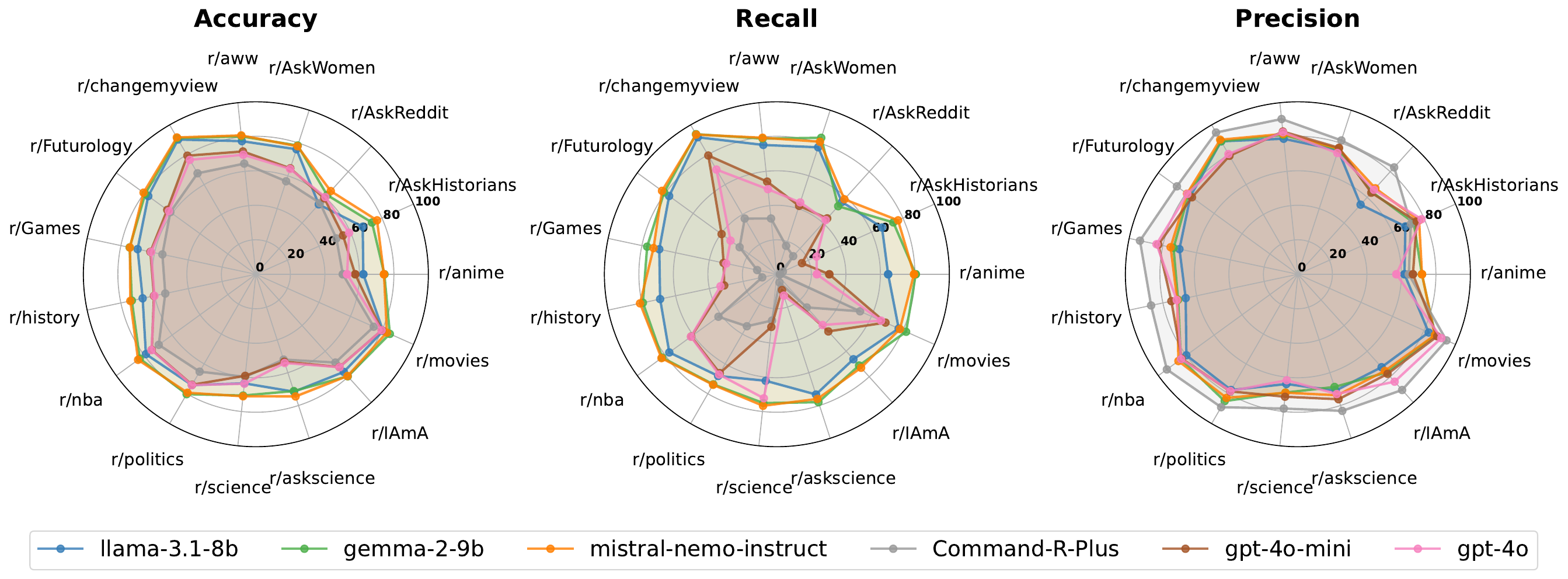} 
    \caption{\textbf{In-domain Moderation Performance.} Comparing the performance of SLMs versus LLMs on accuracy, recall, and precision for in-domain content moderation performance. Best performing SLMs outperform LLMs on accuracy and recall across all subreddits, while LLMs outperform SLMs on precision.\vspace{-12pt}}
    \label{fig:in_domain_results}
\end{figure*}

\subsection{Task}

The task for the language models is to determine a moderation outcome for a given comment when provided with the comment, its preceding context, and subreddit rules. 
For a language model $\mathcal{M}$ and a comment $\mathcal{T}$ along with its context $\mathcal{C}$, and subreddit rules $\mathcal{R},$ we prompt the model with a prompt $p_k$ where $k \in \{0, 2, 4\}$ represents the number of in-context examples provided to the model. Consequently, the moderation task is:
\begin{equation}
    \mathcal{D}_{\mathcal{M}} \leftarrow \mathcal{M}_{p_k}(\mathcal{T}, \mathcal{C}, \mathcal{R})
\end{equation}
where $\mathcal{D}_{\mathcal{M}}$ represents the moderation decision generated by the language model $\mathcal{M}$.

We define two sets of tasks: (a) \textbf{In-domain Moderation}, which involves moderating comments from the same subreddit on which the model was fine-tuned; (b) \textbf{Cross-domain Moderation}, which involves moderating comments from a subreddit different from the one on which the model was fine-tuned. 
Note that this distinction is primarily useful for SLMs. 
For off-the-shelf LLMs, the prompt is subreddit-specific but the models themselves are not, so we always prompt with the correct rules. Furthermore, for the LLMs we use both a zero-shot and few-shot evaluation setup to provide off-the-shelf LLM with some contextual knowledge about the content moderation task.

\subsection{Evaluation Data Distributions}

In real-world moderation settings, the distribution of moderated and unmoderated comments is often heavily skewed. Platforms typically do not report the rates of comments moderated as it ties in with the public perception of the platform. However, \citet{park_measuring_2022} estimated the prevalence of anti-social behavior in 97 popular subreddits at around 5\% (6.25\% in 2016 and 4.28\% in 2020). As a result, we evaluate the performance of models on two kind of data distributions: (a) \textbf{Balanced}, where the number of moderated and unmoderated cases in the test-set are equal; (b) \textbf{Imbalanced}, where the number of moderated comments is 1\%, 5\%, or 10\% of the total number of comments in the test-set.

\subsection{Prompts}

We use rule-based prompting, drawing on prior work~\cite{kumar2024watch}, by querying the models with text, context (preceding comment), and rules from each subreddit using the following prompt $p$:

\begin{table}[ht]
\footnotesize
\centering
\sffamily
\begin{tabular}{p{0.99\columnwidth}}
\rowcollight You are acting as a moderator for the r/\{SUBREDDIT\} subreddit. You will be given a comment from Reddit and the rules deemed suitable to arrive at a moderation outcome, and your task is to determine if the given text is undesirable or not based on the information provided to you.\\
\rowcollight Here is a comment from a Reddit conversation thread, the context (preceding comment), and the associated subreddit rules.\\
\rowcollight \#\#\# Text: \{TEXT\}\\
\rowcollight \#\#\# Context: \{CONTEXT\}\\
\rowcollight \#\#\# Rules:
    \{RULES\}\\
\rowcollight Determine whether the provided text is undesirable or not. Answer with `True' or `False'.\\
\rowcollight \#\#\# Your Response:
\end{tabular}
\end{table}

We obtained rules for each subreddit by querying Reddit using the PRAW API.\footnote{\href{https://praw.readthedocs.io/en/stable/}{https://praw.readthedocs.io/en/stable/}} For in-domain tasks, we use the rules of the original subreddit, whereas, for cross-domain tasks, we use the rules of the source subreddit since we assume that we do not have rules for the target subreddit.

\subsection{Evaluation Metrics}

For our evaluation tasks in the balanced setting, we focus on the metrics of \textit{accuracy}, \textit{precision}, and \textit{recall}. These give us a holistic picture of the efficacy of different models on content moderation tasks, along with an insight into how different models handle violating comments in terms of the precision/recall trade-off. However for the imbalanced setting, we focus on the AUC score as it is a better representation of model performance being less sensitive to class distribution imbalances.

\section{In-domain Moderation}

\subsection{Performance Comparison}\label{sec:in-domain-performance}

In this section, we compare the performance of fine-tuned SLMs versus zero-shot LLMs on in-domain content moderation tasks in a balanced evaluation setting. \autoref{fig:in_domain_results} provides a visual depiction of the accuracy, recall, and precision across models.

\paragraph{Accuracy:}
We evaluate the content moderation performance of each model by comparing it to human moderators' decisions. 
We find that the fine-tuned SLMs\footnote{For detailed performance of base SLMs, see \autoref{app:base_model_performance}.} outperform the LLMs by an average 11.5\% in accuracy across 15 subreddits.
Among the fine-tuned SLMs, Mistral-NeMo-instruct, Gemma-2-9b, and Llama-3.1-8b show accuracies of 77.87\%, 77.2\%, and 72.5\% respectively. Among LLMs, GPT-4o and GPT-4o-mini are the highest-performing models, both with an average accuracy of 65.8\%. SLMs' performance is consistently superior, exhibiting higher accuracy than the LLMs across all subreddits. The highest and lowest average accuracy is achieved by the  SLMs on \textit{r/changemyview} (90.97\%) and \textit{r/AskReddit} (60.4\%), and by the LLMs on \textit{r/movies} (77.4\%) and \textit{r/askscience} (53.4\%).

\paragraph{Precision/Recall Trade-off:}
We find that SLMs demonstrate the highest recall in all 15 subreddits, having an average recall of 77.7\% for Mistral, 77.5\% for Gemma, and 71.5\% for Llama. On the other hand, LLMs achieve the highest precision in 14 out of 15 subreddits, with the highest average precision being for Command R+ at 85.5\%. When comparing the top performers in both metrics, LLMs show an average 8\% advantage in precision, meaning they are more accurate in identifying non-harmful content and avoiding false positives. 
On the other hand, SLMs have a 22\% lead in recall, indicating their superior ability to identify harmful content and reduce false negatives. 
This contrast reveals a notable trade-off: \textit{LLMs excel in minimizing over-flagging of content, whereas SLMs prioritize flagging harmful content even at the expense of more false positives}.

\subsection{Error Analysis: A case study on 
\textit{r/changemyview}}\label{sec:in-domain-analysis} 

In order to get a deeper understanding of how SLMs tackle content moderation and where they falter, we complement our quantitative results with a qualitative analysis. However, due to the scale and complexity of the data, qualitative analysis of the entire dataset is infeasible. Therefore, we focus on one specific community, \textit{r/changemyview} that has been a community of interest in various prior content moderation works~\cite{srinivasan2019content, koshy2023measuring, jhaver2017designing}. \textit{r/changemyview} is a subreddit with 3.7M subscribers for debating opinions where users invite others to challenge their perspectives with thoughtful counterarguments, and if their view is changed, then they can award a \textit{delta} ($\Delta$) to a commenter. Moreover, our fine-tuned SLMs showed strong overall performance on this subreddit, achieving an average of 91\% accuracy, 90\% precision, and 93\% recall, so it would be valuable to further examine the few niche error cases to conduct additional exploration. These factors make it an interesting subreddit to qualitatively compare the differences between content moderation with SLMs versus LLMs, where looking at the false positives and negatives offers valuable insights into the types of errors made by the models.

We retrieved all comments from the test set where the SLMs made an error and inspected them both manually and computationally. Overall, there were 15.2\% (152/1000) false positives and 11.5\% false negatives (115/1000) in the moderation outcomes where at least one SLMs made a mistake.

\paragraph{Impact of Content Length:}

\begin{figure}[t] 
    \centering
    \includegraphics[width=1\linewidth]{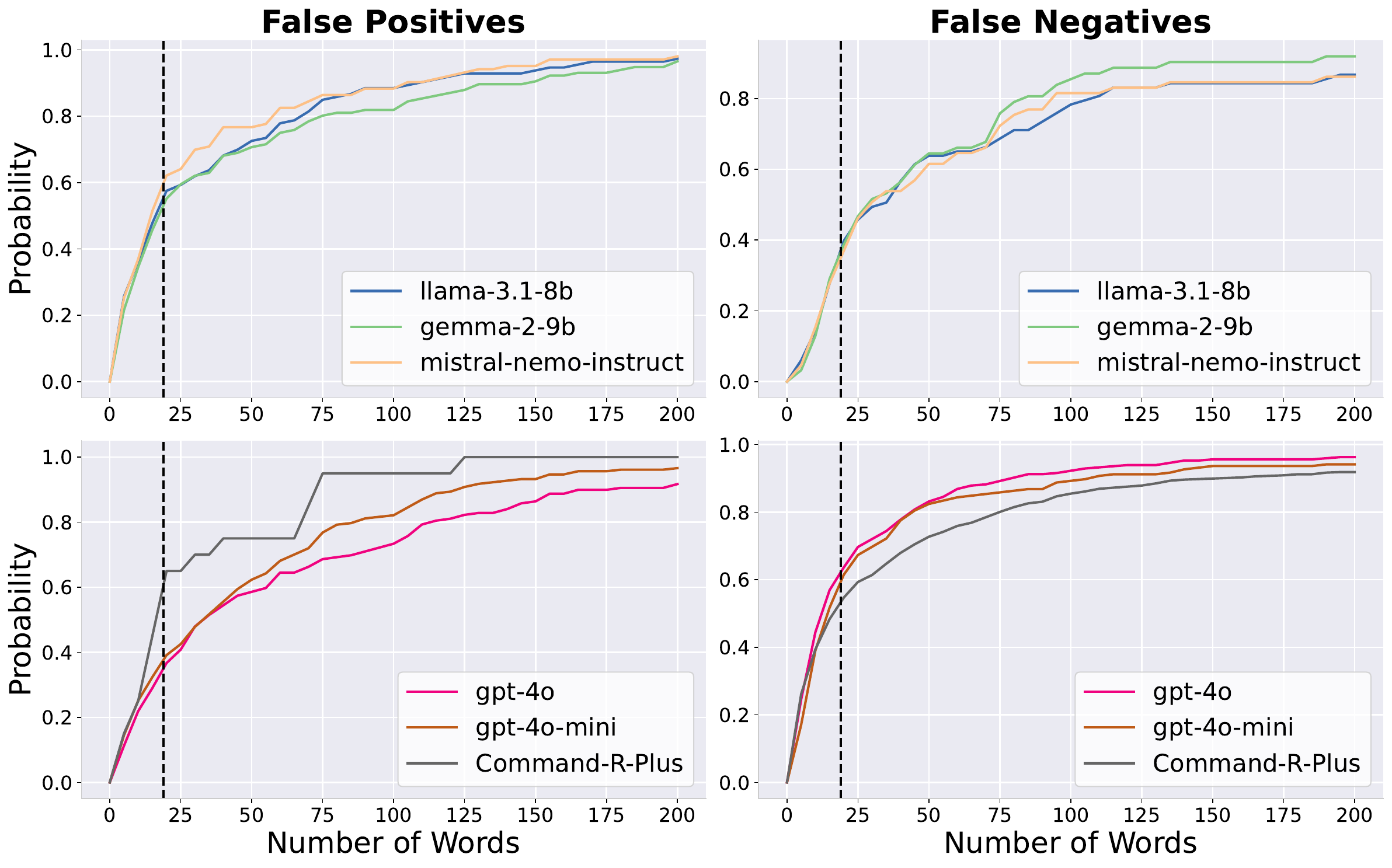} 
    \caption{\textbf{Impact of content length.} Probabilities of the mistakes (FP and FN) made by SLMs and LLMs on varying comment length (in words) in \textit{r/changemyview} reveals that SLMs tend to over-moderate shorter comments whereas LLMs are more forgiving for the same. The vertical bar indicates median length of comments in \textit{r/changemyview} at 19 words.}
    \label{fig:content_length}
\end{figure}

\autoref{fig:content_length} depicts the probability of occurrences of false positive and false negative instances as the number of words in the comment increases. For false positives, we observe that at median comment length, the probability of an SLM moderating the comment incorrectly is around 0.6, while for GPT-4o and GPT-4o-mini, this probability is only 0.4. Command R+ is an exception to the case of LLMs with a probability around 0.65. This means that SLMs are likely to make mistakes and moderate short comments more aggressively compared to LLMs. Qualitatively, we also observe that shorter comments seem to confuse the models more often in terms of false positives. For example, a comment \textit{``Very succinctly stated''}, which was in reply to a comment that has more than 300 words was incorrectly moderated by the SLM, and while a human moderator would understand that this comment is probably said in jest, the SLM might confuse these comments with violating a rule like \textit{``Don't be rude or hostile to other users''}. LLMs, on the other hand, are better suited at handling these kinds of comments and are able to perceive the intended meaning, as all three LLMs correctly left this comment unmoderated.

On the other hand, looking at the false negatives provides us with an opposite observation. The probability of an SLM getting a false negative at median comment length is around 0.4 while for LLMs this probability is around 0.6, with Command R+ having a slightly lower probability. This indicates that short and undesirable comments are well moderated by the SLMs, but are missed by LLMs at a much higher rate.

In both cases, we see that Command R+ performs a bit more like SLMs versus LLMs. While differences in the training data could play a role, it is possible that the model size plays a role in content moderation performance, and with 104B parameters Command R+ can be perceived as being closer to SLMs than to LLMs.

\paragraph{Impact of Content Topic:}

Next, we investigate the impact of content topics that might cause a difference in the performance of the SLMs and LLMs. In order to do so, we perform Latent Dirichlet Allocation (LDA)~\cite{blei2003latent} to extract topics from content in the FPs and FNs for the SLMs. Due to the small number of comments under consideration, we only extract four topics, and follow it up with manual assessment of the comments.

For FPs, three interesting topics we found were (1) web-links in the comments, (2) short comments tagging the comment in the context to respond to a specific point, and (3) comments mentioning \textit{delta} (e.g., \textit{``sounds like you owe him (or me) a delta'', ``$\Delta$''}). Upon inspecting the rules, we found rules that might confuse the model in each of these three instances: (1) \textit{``Doesn't Contribute Meaningfully''} and \textit{``No Neutral/Transgender/Harm a specific person/\textbf{Promo}/Meta''}, (2)  \textit{``No Rude/Hostile Comment''}, and (3) \textit{``No Delta Abuse/Misuse or Should Award Delta''}.

We find that 13.7\% of the FP comments (10/73) contained links, and while all three SLMs got these incorrectly, at least one LLM got each of these 10 moderation outcomes right. Similarly for comments tagging and replying to the previous comment, we find that LLMs correctly identify 51\% (36/73) comments as non-violating. 

On the other hand, for comments mentioning \textit{delta} we actually found that there were some comments in the training set nearly identical to the examples that the SLMs moderated (e.g., ``$\Delta$'', ``!delta rip'', ``Then delta this mannn'') which were removed by human moderators. Further, the rule \textit{``No Delta Abuse/Misuse or Should Award Delta''} is a clear dismissal of these kind of comments. Therefore, our judgment in this case was that the SLMs were correct at moderating them, and perhaps these comments were missed by human moderators.

For FNs, we found that there were many comments related to topics such as wars, political controversies, and gender fluidity which the LLMs always moderate correctly, whereas the SLMs do not. This might be an expected outcome, as all the LLMs in our study have undergone extensive RLHF~\cite{ouyang2022training} and are therefore more cautious when it comes to controversial and sensitive topics, whereas the SLMs are much smaller and do not have the same safety standards. 

\subsection{Does In-Context Learning Improve the Performance of LLMs?}

\begin{table}[t]
\sffamily
\small
\caption{\textbf{In-Context Learning Performance of LLMs:} LLM performance comparison across subreddits for 2-- and 4--shot ICL. Colored by improvement over 0--shot LLM baselines in Section \ref{sec:in-domain-performance}: \colorbox{pink!50}{Reduction}, \colorbox{green!25}{Improvement by $<4\%$}, \colorbox{ForestGreen!50}{Improvement by $\ge4\%$}.}
\label{tab:few-shot-llms}
\resizebox{\columnwidth}{!}{
\begin{tabular}{l|cc|cc|cc}
\textbf{Subreddit~\textbackslash~Model} & \multicolumn{2}{c|}{\textbf{GPT-4o-mini}} & \multicolumn{2}{c|}{\textbf{GPT-4o}} & \multicolumn{2}{c}{\textbf{Command R+}} \\
\hline
\textbf{n--shot} & $n=2$ & $n=4$ & $n=2$ & $n=4$ & $n=2$ & $n=4$ \\
\hline
r/nba & \cellcolor{pink!50}73.1 & \cellcolor{pink!50}74.2 & \cellcolor{green!25}75.2 & \cellcolor{green!25}75.8 & \cellcolor{ForestGreen!50}73.3 & \cellcolor{ForestGreen!50}74.7 \\
r/aww & \cellcolor{pink!50}65.7 & \cellcolor{pink!50}70.8 & \cellcolor{green!25}70.7 & \cellcolor{green!25}70.2 & \cellcolor{ForestGreen!50}70.1 & \cellcolor{ForestGreen!50}70.3 \\
r/movies & \cellcolor{pink!50}71.7 & \cellcolor{pink!50}77.2 & \cellcolor{green!25}80.8 & \cellcolor{pink!50}79.1 & \cellcolor{ForestGreen!50}77.8 & \cellcolor{green!25}76.9 \\
r/politics & \cellcolor{pink!50}62.1 & \cellcolor{pink!50}69.8 & \cellcolor{green!25}74.7 & \cellcolor{green!25}75.8 & \cellcolor{ForestGreen!50}69.8 & \cellcolor{ForestGreen!50}70.3 \\
r/changemyview & \cellcolor{pink!50}68.8 & \cellcolor{pink!50}71.8 & \cellcolor{pink!50}75.5 & \cellcolor{pink!50}76.2 & \cellcolor{ForestGreen!50}74.6 & \cellcolor{ForestGreen!50}71.1 \\
\hline
\end{tabular}
}
\end{table}

Since we have used LLMs in a zero-shot manner so far, we investigated whether providing examples to the LLM for in-context learning (ICL) could improve their performance. To do so, we use $n-$shot ICL ($n\in\{2, 4\}$) and compare to the zero-shot setup in \autoref{tab:few-shot-llms} for the five subreddits where the LLMs showed the highest performance. In order to mitigate potential biases in the chosen examples for ICL, we randomly sampled $2$ or $4$ examples from the training split for each setting.

\begin{figure}[t] 
    \centering
    \includegraphics[width=1\linewidth]{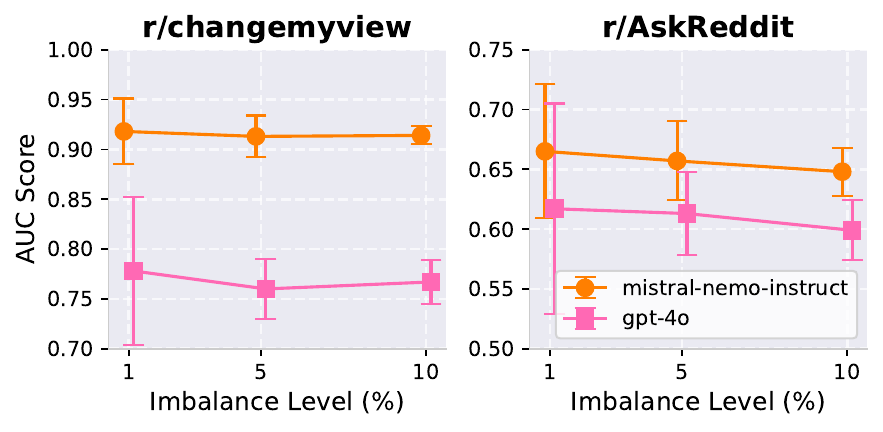} 
    \caption{\textbf{Imbalanced Distribution Evaluation.} Best performing SLM (Mistral-NeMo-Instruct) and LLM (GPT-4o) on $1\%$, $5\%$, and $10\%$ imbalance-level test split of \textit{r/changemyview} and \textit{r/AskReddit} by AUC scores. Error bars represent standard deviation over 30 seeds.\vspace{-15pt}}
    \label{fig:imbalanced_auc}
\end{figure}

We see that ICL either leads to a \colorbox{pink!50}{reduction} in performance of the LLMs or provides marginal gains (\colorbox{green!25}{$< 4\%$}) for both GPT-4o-mini and GPT-4o. For Command R+ we see a performance gain of \colorbox{ForestGreen!50}{over $4\%$} in most cases, but is still outperformed by GPT-4o. Overall, we observe that including ICL examples for LLMs in content moderation task can be unstable and fails to help LLMs match the performance of SLMs. Our results match those of \citet{guo2023investigation} on the related task of hate-speech detection where few-shot prompting leads to the lowest performance across various prompting techniques used by the researchers.

\subsection{How Does Imbalanced Data Affect the Performance of SLMs versus LLMs?}

We now evaluate the content moderation performance of language models on imbalanced datasets, as balanced sampling results in a less challenging distribution than real-time moderation data where the number of removals are typically low ($< 10\%$).

We compute the test metrics at different imbalance thresholds: 1\%, 5\%, and 10\% moderated comments and remaining unmoderated comments. Due to the possibility of high variance based on the sample, we collate AUC scores for 30 different runs. We report results here for 2 subreddits---\textit{r/changemyview} and \textit{r/AskReddit}---where the SLMs had the best and the worst performance, respectively, in the balanced setting. We pick the SLM and LLM with the best performance at each imbalance level, which were Mistral-NeMo-Instruct and GPT-4o in all cases. 

From \autoref{fig:imbalanced_auc} we notice that even under an imbalanced distribution, Mistral-Nemo-Instruct outperforms GPT-4o on moderating content from both subreddits with average AUC of 0.7915, 0.785, and 0.781
compared to the average AUC of 0.698, 0.687, 0.683 of GPT-4o at a 1\%, 5\%, and 10\% imbalance level respecively.

\begin{figure*}[t] 
    \centering
    \includegraphics[width=1\textwidth]{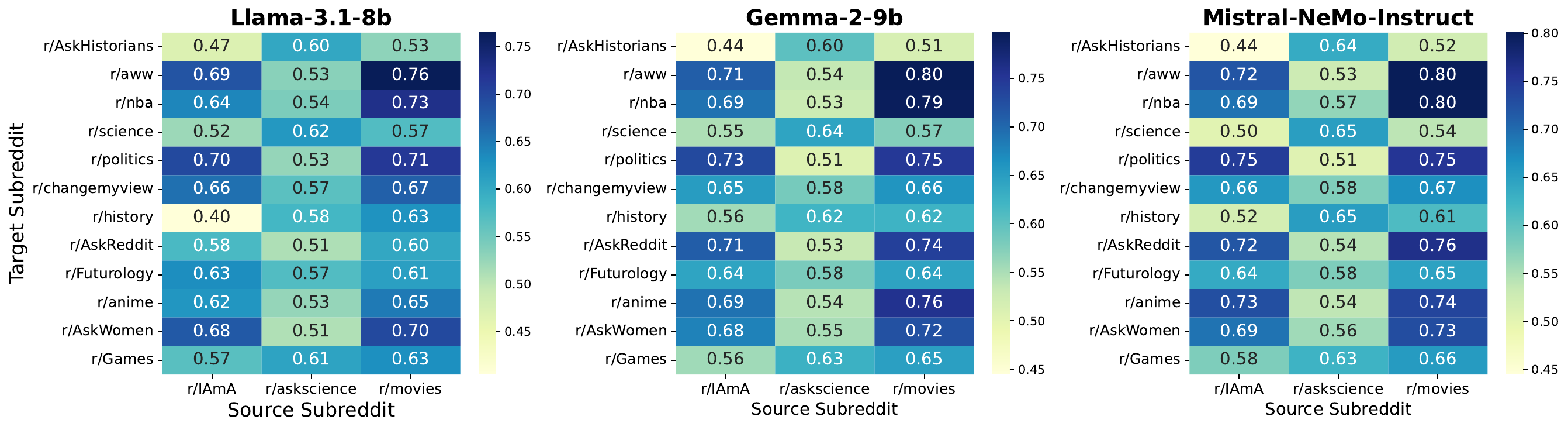} 
    \caption{\textbf{Cross-domain Moderation Performance.} Comparison of performance of SLMs in terms of accuracy for cross-domain content moderation performance on three target subreddits: \textit{r/IAmA}, \textit{r/askscience}, \textit{r/movies}. Mistral-NeMo-Instruct gives the best cross-domain performance, with 75\% accuracy for \textit{r/IAmA} by the model fine-tuned for \textit{r/politics}, 65\% accuracy for \textit{r/askscience} by the model fine-tuned for \textit{r/science} and \textit{r/history}, and 80\% for \textit{r/movies} by the model fine-tuned for \textit{r/aww} and \textit{r/nba}.}
    \label{fig:cross_domain_results}
\end{figure*}

\section{Cross-domain Moderation}
\subsection{Performance Comparison}

In this section, we investigate the cross-domain performance of SLMs on moderation tasks.~\autoref{fig:cross_domain_results} shows a visual depiction of the findings.
We choose three communities examined in prior work, namely \textit{r/IAmA}, \textit{r/askscience}, and \textit{r/movies} as our three subreddits for the discussion of cross-domain performance~\cite{kumar2024watch}.~\autoref{app:cross_domain_detailed} provides complete cross-domain results.

We observe that \textit{r/askscience} is the subreddit with the lowest average cross-domain performance at 58\% with the highest performance of 65\% given by the Mistral fine-tuned for \textit{r/science} and \textit{r/history}. On the other hand, \textit{r/movies} was the easiest of the three subreddits, with the highest average cross-domain performance at 68.6\% and a highest performance of 80\% given by the Mistral fine-tuned for \textit{r/aww} and \textit{r/nba}, and Gemma model fine-tuned for \textit{r/aww}. Finally, for \textit{r/IAmA}, the average accuracy was 63.7\% and the highest performance came from the Mistral model fine-tuned for \textit{r/politics} at 75\%.

It is noteworthy that while the moderation accuracy of cross-domain models is not at par with in-domain SLM models, the best-performing cross-domain SLMs outperform best-performing LLMs for these subreddits, with relative advantages of 6.5\% for \textit{r/IAmA}, 11.6\% for \textit{r/askscience}, and 0.1\% for \textit{r/movies}, indicating the promise of cross-community based approach for content moderation using SLMs over using LLMs.

Furthermore, we see that there are some models that show a promising overall cross-domain performance. Specifically, models trained on \textit{r/changemyview}, \textit{r/nba}, and \textit{r/movies} show an average cross-domain accuracy between 64.6\% and 67.8\%, and can therefore be used as \textit{`meta-experts'} to moderate communities that would benefit more from a cross-domain moderation as compared to in-domain moderation. One example is \textit{r/AskReddit}, which would benefit from the cross-domain models from 10 other communities compared to its in-domain model based on Llama-3.1-8b and Gemma-2-9b, and 7 others for Mistral-NeMo-Instruct.

\subsection{Exploring factors that affect cross-domain performance of SLMs}

Due to the promising performance of cross-domain models, we further investigate why some subreddits may benefit from cross-domain moderation, and whether there is a pattern to the cross-domain accuracy of an SLM on a target subreddit by testing the impact of subreddit size, description and rules. Our hypothesis is that relative sizes of subreddits can play a role in determining content moderation performance across communities as similarly-sized communities may have similar underlying norms. Accordingly, the similarity of topics as well as community rules between subreddits may have a direct impact on cross-community content moderation outcomes, due to the similarity in content or in the way in which the content is moderated.

For size, we construct a matrix of relative number of subscribers of the source subreddit w.r.t to the target subreddit, and for description and rules, we get the subreddit description and rules using PRAW and use Cohere's \texttt{embed-english-v3.0}~\cite{cohereIntroducingEmbed} to generate embeddings and compute two cosine-similarity matrices between pairs of source and target subreddit descriptions and rules.

We then conduct a column-wise (in this case, subreddit-wise) t--test for non-correlation between the relative-size matrix, and the cosine similarity matrices for description and rules, with the cross-domain accuracy matrices for each model, using the Pearson correlation coefficient~\cite{schober2018correlation} as our test-statistic ($r$). The alternative hypothesis was that there exists a positive correlation.

We find that there was no statistically significant ($\alpha = 0.05$) positive correlation between relative size and the performance of cross-domain models on a subreddit. For subreddit description, we find two statistically significant positive correlations with Llama-3.1-8b on \textit{r/AskHistorians} ($t(12)=0.459, p=.049$) and
Gemma-2-9b on \textit{r/askscience} ($t(12)=0.535, p=.024$). Finally, for correlation between subreddit rules and cross-domain performance we again find only two significant positive correlations with Llama-3.1-8b on \textit{r/nba} ($t(12)=0.462, p=.048$) and
Gemma-2-9b on \textit{r/anime} ($t(12)=0.459, p=.049$). Full results of the t-test can be found in \autoref{app:cross_domain_corr_test}.

This result signifies that while the topic of the subreddit and the community rules may play some role in determining the cross-domain performance, the overall notions of what determines the cross-community performance go beyond just subreddit sizes, description, and rules.

\section{Discussion and Implications}

In this section, we discuss the implications of our work for online content moderation.

\subsection{Shift from generalist LLMs to specialist SLMs for content moderation}

Our findings indicate a significant advantage in employing SLMs for content moderation tasks over LLMs, consistently showing superior performance in terms of both accuracy and recall. 
Our findings suggest that SLMs adopt a more aggressive approach compared to LLMs, resulting in significantly higher recall for SLMs, though at the cost of slightly lower precision relative to LLMs. However, prior work has shown that moderators, who are overburdened already, may not be able to attend to all instances of norm-violating and undesirable behavior~\cite{park2022measuring, chandrasekharan_crossmod_2019}. Hence, the higher recall of SLMs compared to LLMs could actually provide a benefit for more reliable detection of harmful content that could otherwise be overlooked and stay on the platform for longer---potentially leading to further undesirable outcomes~\cite{lambert2022conversational}.

Apart from performance, a key strength of SLMs lies in their ability to be fine-tuned for specific communities, allowing for moderation that aligns closely with the unique needs of individual subreddits. 
Additionally, specialized communities and those serving sensitive populations~\cite{saha2020understanding} may require extra safeguards and considerations when using automated content moderation tools, which can be more easily fine-tuned with SLMs.
This approach improves moderation accuracy by better reflecting the decisions of community-specific moderators. 
In addition, community norms can evolve over time, and performing continual pretraining~\cite{ke2023continual} on SLMs with incrementally collected data can help the models stay updated in accordance. 

Finally, SLMs for content moderation are a cheaper and more scalable option for platforms like Reddit, which manage large volumes of user-generated content and provide them agency over their models and data without the reliance on third-party APIs. LLMs, on the other hand, are expensive to query, rate-limited, and mostly closed-source.

\subsection{Automated tools for removals vs. triaging}

The precision/recall trade-off between LLMs and SLMs discussed in the previous section also has implications for their potential usage from a moderation design perspective. Specifically, content moderation can either be done in an automated manner with no moderator involvement or in a moderator-in-the-loop manner where triaged and reported comments are manually moderated.

Since LLMs are better at accurately identifying violating comments while minimizing false positives, they are more suited to be used as automated moderation tools as they are less likely to wrongfully penalize community members by moderating potentially benign comments due to their cautious approach. SLMs, on the other hand, are more aggressive at flagging comments with higher recall, which makes them more suited to scenarios where the priority is to ensure that potentially harmful content is quickly triaged and sent for further review by human moderators. This will ensure the flagging of seemingly undesirable behavior in a time-sensitive manner and not leave such content visible on the platform for long periods, and upon manual inspection, if the comment is benign, it can be allowed to exist by the moderator.

\subsection{Cross-domain approaches to moderation}

We observe that along with great in-domain performance, SLMs also perform well in cross-domain settings which suggests that norm violation representations learned by fine-tuned SLMs can generalize effectively across different online communities, making them a viable option for moderating content in new or growing communities. 

Moreover, we saw that certain \textit{`meta-expert'} models have high average cross-domain performance, often providing certain communities with more performance than their in-domain models. This indicates that SLMs have the capability to learn shared norm violation representations, and by leveraging cross-community similarities, cross-domain SLMs can be used for moderation without needing specific training data for every new community, thus enabling faster and more cost-efficient deployment of automated moderation tools across platforms. Specifically, a new community developing its rules can make use of cross-domain experts to identify which expert provides the community with the highest content moderation performance, and use that to inform their own community rules.

However, we show that determining which cross-domain expert would provide a community with the highest benefit is a nuanced task that goes beyond measures like similarity in subreddit sizes, descriptions, and rules. This provides an avenue for future research to explore the development of strategies to identify the right set of cross-community experts as well as meta-moderation models. Using a mixture of experts (MoE) framework, for example, could allow a community to draw from specialized models trained on different clusters of subreddits with shared themes, rules, or behaviors. This would enable automated moderation systems to dynamically route content to the most appropriate expert and ensure that moderation decisions are well-aligned with specific community norms.

\subsection{Instability of Closed-source LLMs}

Since closed-source models undergo frequent updates, it is crucial to assess the stability of their performance in content moderation tasks. To analyze how different releases of the same model can affect performance, we compare the initial release of OpenAI's GPT-4o model (May 13, 2024) with the latest version (August 06, 2024). 

From \autoref{tab:gpt-comparison}, we observe that the accuracy of GPT-4o declined from 67.7\% to 65.8\%, accompanied by a huge reduction in recall from 55.7\% to 44.7\%. Conversely, the average precision improved from 73.6\% to 76.3\%. Crucially, 10.7\% or 2,166 comments across all subreddits that were correctly moderated by the May release of the model, were no longer moderated by the August version, highlighting a lack of trust that could be placed in these closed-source LLMs for content moderation.

Given this black-box nature of closed-source models, ensuring consistent moderation performance would pose a challenge when using closed-source models. Moderators may instead consider using open-source language models, which tend to be more stable since moderators can choose not to update the models and can be fine-tuned more easily to meet specific moderation requirements.

\begin{table}[t]
\sffamily
\small
\caption{\textbf{Instability of GPT-4o.} Average performance metrics for two different releases of GPT-4o.}
\label{tab:gpt-comparison}
\resizebox{\columnwidth}{!}{
\begin{tabular}{lcc}
\textbf{Metric} & \textbf{GPT-4o (2024-05-13)} & \textbf{GPT-4o (2024-08-06)}\\
\midrule
Precision & 0.736 & \textbf{0.763} \\
Recall & \textbf{0.557} & 0.447 \\
Accuracy & \textbf{0.677} & 0.658\\
\bottomrule
\end{tabular}
}
\end{table}

\section{Conclusion}

This paper examined the effectiveness of small language models (SLMs) and large language models (LLMs) in content moderation tasks. We found that SLMs with less than 15b parameters, such as Gemma-2-9b and Mistral-Nemo-Instruct, consistently outperformed significantly larger LLMs like GPT-4o at identifying undesirable content. SLMs showed a higher recall, indicating their superior ability to flag a wider range of undesirable content, which can be crucial for effective moderation on large-scale platforms. We also found that SLMs have higher AUC scores than LLMs under realistic imbalanced data conditions, and that even with in-context examples LLMs fail to match SLMs. In addition to in-domain moderation tasks, we uncovered the potential of SLMs to be used in cross-community moderation tasks. Cross-community moderation can benefit smaller communities with fewer resources to train their own in-domain models or newer communities where community norms are still emerging. We provided qualitative insights into the trade-offs arising from content length and topic that could impact SLMs and LLMs differently. Overall, SLMs offer an effective, scalable, and cost-effective solution for content moderation, achieving a strong balance between accuracy, recall, and stability. Future work can further improve cross-domain performance by training models that learn shared notions of norms and values, and explore frameworks that leverage complementary strengths of SLMs and LLMs to enhance moderation capabilities across diverse online communities.
\section{Limitations}

\subsection{Scale of Subreddit Selection}

We chose 15 highly popular subreddits for our study with a good mix of science, AMA, entertainment, history, sports, and political subreddits in order to capture a wide spectrum of underlying norms and subreddit characteristics. We therefore believe that our findings are representative and scalable to a wider range of communities. However, future studies could expand the scope to incorporate a larger number of subreddits for increased robustness of our findings and more open-sourced models for content moderation.

\subsection{Text-Based Content Moderation}

Due to availability of publicly available text-based datasets for Reddit comment removals, we resorted to purely text-based content moderation. However, undesirable behavior may occur in multiple other modes like images and future studies can expand on our insights to explore the performance of vision-language models in content moderation settings.

\subsection{Continual Updating of Models}

Community norms and notions of undesirable behavior may change over time and models fine-tuned on a specific date range of posting activity may not necessarily generalize to newer comment removals or an evolved representation of content that moderators would remove. Therefore, future studies can evaluate the adaptability of models on temporal distribution shifts in online content and determine the efficacy of techniques like domain-adaptive pretraining (DAPT)~\cite{gururangan2020don} and continual pretraining~\cite{ke2023continual} in keeping SLMs update with the latest norms.

\section*{Ethical Considerations}

Our work explores the use of language models for online content moderation and shows the promise of small-scale fine-tuned models to achieve superior performance. While small language models provide freedom to fine-tune the models as required, this can potentially have consequences if used by communities and moderators in an adversarial manner to moderate comments from particular users or factions of society therefore leading to unintended ethical consequences in terms of freedom of rightful expression on social media platforms. Finally, since we work with the OpenAI\footnote{\href{https://openai.com/policies/terms-of-use/}{https://openai.com/policies/terms-of-use/}} and Cohere\footnote{\href{https://cohere.com/terms-of-use}{https://cohere.com/terms-of-use}} APIs, we ensure to comply with their terms of use policies.

\section*{Acknowledgments}

The authors sincerely thank the anonymous reviewers for their constructive feedback on our work during the ARR peer review stage. A.G. was supported by compute credits from a Cohere For AI Research Grant. This work used the Delta system at the National Center for Supercomputing Applications through allocation \#240481 from the Advanced Cyberinfrastructure Coordination Ecosystem: Services \& Support (ACCESS) program, which is supported by National Science Foundation grants \#2138259, \#2138286, \#2138307, \#2137603, and \#2138296.

\bibliography{references}

\appendix

\section{Subreddits Statistics}\label{app:subreddits_stats}

In \autoref{tab:subreddit_size_description} we list the number of subscribers in each of the subreddits we stufy in our work along with thei rofficial public description from Reddit.

\begin{table*}[t]
\sffamily
\small
\caption{Base Small Language Model (SLM) Metrics by Subreddit.}
\label{tab:base_slm_results}
\resizebox{\linewidth}{!}{
\begin{tabular}{l|ccc|ccc|ccc}
& \multicolumn{3}{c|}{\textbf{llama-3.1-8b}} & \multicolumn{3}{c|}{\textbf{gemma-2-9b}} & \multicolumn{3}{c}{\textbf{mistral-nemo-instruct}} \\
\hline
\textbf{Subreddit} & \textbf{Precision} & \textbf{Recall} & \textbf{Accuracy} & \textbf{Precision} & \textbf{Recall} & \textbf{Accuracy} & \textbf{Precision} & \textbf{Recall} & \textbf{Accuracy} \\
\hline
\textbf{r/AskHistorians} & 0.511 & 0.594 & 0.513 & 0.000 & 0.000 & 0.500 & 0.772 & 0.044 & \textbf{0.516} \\
\textbf{r/AskReddit} & 0.514 & 0.622 & 0.516 & 0.846 & 0.011 & 0.504 & 0.701 & 0.286 & \textbf{0.582} \\
\textbf{r/AskWomen} & 0.498 & 0.600 & 0.497 & 0.667 & 0.004 & 0.501 & 0.820 & 0.201 & \textbf{0.578} \\
\textbf{r/Futurology} & 0.507 & 0.605 & 0.508 & 0.500 & 0.005 & 0.500 & 0.800 & 0.308 & \textbf{0.616} \\
\textbf{r/Games} & 0.495 & 0.634 & 0.494 & 0.656 & 0.040 & 0.509 & 0.868 & 0.132 & \textbf{0.556} \\
\textbf{r/IAmA} & 0.398 & 0.654 & 0.460 & 0.353 & 0.009 & 0.592 & 0.816 & 0.410 & \textbf{0.724} \\
\textbf{r/anime} & 0.500 & 0.652 & 0.500 & 0.500 & 0.004 & 0.500 & 0.819 & 0.068 & \textbf{0.526} \\
\textbf{r/askscience} & 0.497 & 0.604 & 0.496 & 0.167 & 0.001 & 0.498 & 0.744 & 0.067 & \textbf{0.522} \\
\textbf{r/aww} & 0.510 & 0.622 & 0.512 & 0.333 & 0.002 & 0.499 & 0.875 & 0.371 & \textbf{0.659} \\
\textbf{r/changemyview} & 0.512 & 0.661 & 0.515 & 0.400 & 0.004 & 0.499 & 0.861 & 0.408 & \textbf{0.671} \\
\textbf{r/history} & 0.497 & 0.527 & 0.496 & 0.567 & 0.017 & 0.502 & 0.802 & 0.097 & \textbf{0.536} \\
\textbf{r/movies} & 0.490 & 0.630 & 0.486 & 0.556 & 0.005 & 0.500 & 0.912 & 0.560 & \textbf{0.753} \\
\textbf{r/nba} & 0.491 & 0.623 & 0.489 & 0.320 & 0.008 & 0.496 & 0.892 & 0.531 & \textbf{0.734} \\
\textbf{r/politics} & 0.512 & 0.577 & 0.514 & 1.000 & 0.001 & 0.500 & 0.817 & 0.401 & \textbf{0.656} \\
\textbf{r/science} & 0.510 & 0.535 & 0.510 & 0.295 & 0.013 & 0.491 & 0.723 & 0.191 & \textbf{0.559} \\
\hline
\end{tabular}
}
\end{table*}

\begin{table*}[t]
\sffamily
\caption{Subscriber sizes and descriptions of the 15 subreddits studied in this work.}
\label{tab:subreddit_size_description}
\resizebox{1\textwidth}{!}{
\begin{tabular}{lrl}
\hline
\textbf{Subreddit} & \textbf{Size} & \textbf{Subreddit Description} \\
\toprule
r/askscience & 26M & A subreddit for people to ask a science question, and get a science answer \\
r/IAmA  & 23M &  A Q\&A subreddit featuring interactive interviews with individuals of various backgrounds\\
r/movies & 34M & A space for inclusive discussions on films, including reviews and news about major releases\\
r/anime &11M & A subreddit for Reddit's premier anime community\\
r/AskHistorians &2.1M & A subreddit for well-researched, expert-level answers on historical questions\\
r/AskReddit &49M & A subreddit to ask and answer thought-provoking questions\\
r/AskWomen &5.5M & A subreddit for women to share their perspectives and experiences on various topics\\
r/aww &37M & A subreddit for cute and cuddly pictures\\
r/changemyview &3.7M & A subreddit for users to present opinions they are open to having challenged through reasoned debate\\
r/Futurology&21M & A subreddit devoted to the field of Future(s) Studies and evidence-based speculation about the development of humanity, technology, and civilization\\
r/Games &3.3M & A subreddit to provide a place for informative and interesting gaming content and discussions. \\
r/history &18M & A subreddit for discussions about history\\
r/nba &13M & A subreddit for NBA discussion\\
r/politics &8.7M & A subreddit for news and discussion about U.S. politics.\\
r/science &33M & A subreddit to share and discuss new scientific research. \\
\hline
\end{tabular}
}
\end{table*}

\begin{table*}[t]
\sffamily
\caption{Fine-tuned Small Language Model (SLM) and Large Language Model (LLM) Accuracy by Subreddit.}
\resizebox{1\textwidth}{!}{
\begin{tabular}{l|ccc|ccccc}
\textbf{Subreddit} & \textbf{llama-3.1-8b} & \textbf{gemma-2-9b} & \textbf{mistral-nemo-instruct} & \textbf{Command-R-Plus} & \textbf{gpt-4o-mini} & \textbf{gpt-4o} & \textbf{gpt-4o-2024-05-13} & \textbf{gpt3.5-turbo} \\
\hline
r/askscience & 0.716 & 0.712 & \textbf{0.745} & 0.520 & 0.533 & 0.541 & 0.550 & 0.541\\
r/IAmA & 0.762 & \textbf{0.796}& 0.794 & 0.689 & 0.724 & 0.725 & 0.731 & 0.604 \\
r/movies & 0.807 & \textbf{0.850}& 0.832 & 0.748 & 0.799 & 0.798 & 0.809 & 0.752\\
r/anime & 0.623 & \textbf{0.745} & 0.743& 0.504 & 0.576 & 0.529 & 0.574 & 0.576\\
r/AskHistorians  & 0.678 & 0.737& \textbf{0.769} & 0.511 & 0.553 & 0.592 & 0.593 & 0.563\\
r/AskReddit  & 0.545 & 0.618& \textbf{0.649} & 0.556 & 0.593 & 0.602 & 0.598 & 0.577\\
r/AskWomen & 0.762 & \textbf{0.785}& 0.781 & 0.567 & 0.646 & 0.642 & 0.669 & 0.620\\
r/aww & 0.776 & 0.804& \textbf{0.810} & 0.645 & 0.716 & 0.697 & 0.753 & 0.691\\
r/changemyview  & 0.902 & 0.911& \textbf{0.916} & 0.676 & 0.794 & 0.766 & 0.768 & 0.623\\
r/Futurology & 0.771 & 0.793& \textbf{0.805} & 0.613 & 0.635 & 0.623 & 0.646 & 0.649\\
r/Games & 0.701 & \textbf{0.748}& 0.747 & 0.554 & 0.625 & 0.621 & 0.675 & 0.610\\
r/history & 0.671 & 0.735& \textbf{0.744} & 0.537 & 0.604 & 0.601 & 0.630 & 0.597\\
r/nba & 0.788 & 0.832 & \textbf{0.844}& 0.696 & 0.746 & 0.748 & 0.778 & 0.699\\
r/politics  & 0.742 & \textbf{0.803}& 0.793 & 0.652 & 0.740 & 0.742 & 0.736 & 0.692\\
r/science  & 0.636 & 0.707& \textbf{0.712} & 0.598 & 0.592 & 0.638 & 0.641 & 0.561\\
\hline
Avg. & 0.725 & 0.772 & \textbf{0.779}& 0.605 & 0.658 & 0.658 & 0.677 & 0.624\\
\hline
\end{tabular}
}
\end{table*}

\begin{table*}[h]
\centering
\sffamily
\caption{Fine-tuned Small Language Model (SLM) and Large Language Model (LLM) Precision by Subreddit.}
\resizebox{1\textwidth}{!}{
\begin{tabular}{l|ccc|ccccc}
\textbf{Subreddit} & \textbf{llama-3.1-8b} & \textbf{gemma-2-9b}& \textbf{mistral-nemo-instruct} & \textbf{Command-R-Plus} & \textbf{gpt-4o-mini} & \textbf{gpt-4o} & \textbf{gpt-4o-2024-05-13} & \textbf{gpt3.5-turbo}\\
\hline
r/askscience & 0.709 & 0.687 &0.738 & \textbf{0.833} & 0.762 & 0.730 & 0.771 & 0.617\\
r/IAmA  & 0.725 & 0.767&0.754 & \textbf{0.903} & 0.777 & 0.838 & 0.736 & 0.508\\
r/movies & 0.830 & 0.872&0.869 & \textbf{0.943} & 0.884 & 0.910 & 0.855 & 0.759\\
r/anime & 0.618 & 0.719&\textbf{0.720} & 0.643 & 0.667 & 0.571 & 0.640 & 0.579\\
r/AskHistorians  & 0.682 & 0.735&0.769 & 0.717 & 0.754 & \textbf{0.785} & 0.765 & 0.584\\
r/AskReddit  & 0.543 & 0.642&0.671 & \textbf{0.834} & 0.636 & 0.659 & 0.599 & 0.583\\
r/AskWomen & 0.756 & 0.761&0.766 & \textbf{0.816} & 0.771 & 0.739 & 0.708 & 0.633\\
r/aww & 0.790 & 0.812& 0.818& \textbf{0.905} & 0.832 & 0.828 & 0.815 & 0.716\\
r/changemyview  & 0.890 & 0.890&0.901 & \textbf{0.949} & 0.793 & 0.805 & 0.721 & 0.616\\
r/Futurology & 0.770 & 0.784&0.794 & \textbf{0.867} & 0.759 & 0.792 & 0.734 & 0.662\\
r/Games & 0.702 & 0.737&0.754 & \textbf{0.936} & 0.827 & 0.838 & 0.821 & 0.754\\
r/history & 0.665 & 0.710&0.715 & \textbf{0.870} & 0.750 & 0.718 & 0.718 & 0.604\\
r/nba & 0.799 & 0.838& 0.856& \textbf{0.939} & 0.833 & 0.837 & 0.829 & 0.690\\
r/politics  & 0.775 & 0.849& 0.828 & \textbf{0.890} & 0.784 & 0.782 & 0.721 & 0.685\\
r/science  & 0.640 & 0.690& 0.690 & \textbf{0.783} & 0.714 & 0.618 & 0.609 & 0.600\\
\hline
Avg. & 0.726 & 0.766&0.776 & \textbf{0.855} & 0.770 & 0.763 & 0.736 & 0.639\\
\hline
\end{tabular}
}
\end{table*}

\begin{table*}[h]
\sffamily
\caption{Fine-tuned Small Language Model (SLM) and Large Language Model (LLM) Recall by Subreddit.}
\resizebox{1\textwidth}{!}{
\begin{tabular}{l|ccc|ccccc}
\textbf{Subreddit} & \textbf{llama-3.1-8b} & \textbf{gemma-2-9b}& \textbf{mistral-nemo-instruct} & \textbf{Command-R-Plus} & \textbf{gpt-4o-mini} & \textbf{gpt-4o} & \textbf{gpt-4o-2024-05-13} & \textbf{gpt3.5-turbo}\\
\hline
r/askscience & 0.733 & \textbf{0.780}&0.760 & 0.050 & 0.096 & 0.130 & 0.140 & 0.225\\
r/IAmA  & 0.663 & 0.712&\textbf{0.729} & 0.259 & 0.446 & 0.396 & 0.529 & 0.723\\
r/movies & 0.773 & \textbf{0.820}&0.781 & 0.528 & 0.689 & 0.661 & 0.748 & 0.744\\
r/anime & 0.645 & \textbf{0.805}&0.795 & 0.018 & 0.304 & 0.232 & 0.334 & 0.564\\
r/AskHistorians  & 0.667 & 0.739&\textbf{0.769} & 0.038 & 0.159 & 0.252 & 0.274 & 0.448\\
r/AskReddit  & 0.564 & 0.531&\textbf{0.585} & 0.141 & 0.436 & 0.423 & 0.570 & 0.521\\
r/AskWomen & 0.774 & \textbf{0.832}&0.808 & 0.173 & 0.417 & 0.438 & 0.571 & 0.571\\
r/aww & 0.754 & 0.791&\textbf{0.796} & 0.325 & 0.540 & 0.496 & 0.654 & 0.632\\
r/changemyview  & 0.917 & \textbf{0.938}&0.935 & 0.373 & 0.795 & 0.701 & 0.895 & 0.661\\
r/Futurology & 0.772 & 0.808&\textbf{0.823} & 0.267 & 0.396 & 0.332 & 0.497 & 0.612\\
r/Games & 0.697 & \textbf{0.769}&0.731 & 0.117 & 0.316 & 0.300 & 0.450 & 0.330\\
r/history & 0.692 & 0.794&\textbf{0.811} & 0.087 & 0.312 & 0.334 & 0.439 & 0.551\\
r/nba & 0.771 & 0.822 &\textbf{0.827}& 0.419 & 0.614 & 0.616 & 0.702 & 0.717\\
r/politics  & 0.681 & 0.737&\textbf{0.740} & 0.348 & 0.662 & 0.671 & 0.767 & 0.710\\
r/science  & 0.621 & 0.751&\textbf{0.767} & 0.270 & 0.307 & 0.723 & 0.789 & 0.367\\
\hline
Avg. & 0.715 & 0.775 &\textbf{0.777}& 0.228 & 0.433 & 0.447 & 0.557 & 0.558\\
\hline
\end{tabular}
}
\end{table*}

\section{LoRA Hyperparameters}

In order to fine-tune our community-specific LLMs, we perform Low-Rank Adaptation (LoRA)~\cite{hu2021lora} for 1 epoch on 8,000 balanced samples from publicly available content moderation datasets. We use rank $r = 16$ LoRA with an $\alpha = 32$ and no dropout. We use $5$ warmup steps, a linear learning rate schedule with $lr=2e-4$, AdamW~\cite{loshchilov2017decoupled} optimizer with a weight decay of $0.01$.

\section{Compute Resources}

All experiments on open-source models were run on a GPU server equipped with 1xNVIDIA A100. The experiments with the OpenAI models cost about 250 USD and experiments with Cohere Command R+ and Cohere Embedv3 English cost about 30 USD.

\section{Detailed Cross-Domain Results}\label{app:cross_domain_detailed}

\begin{figure*}[ht] 
    \centering
    \includegraphics[width=1\textwidth]{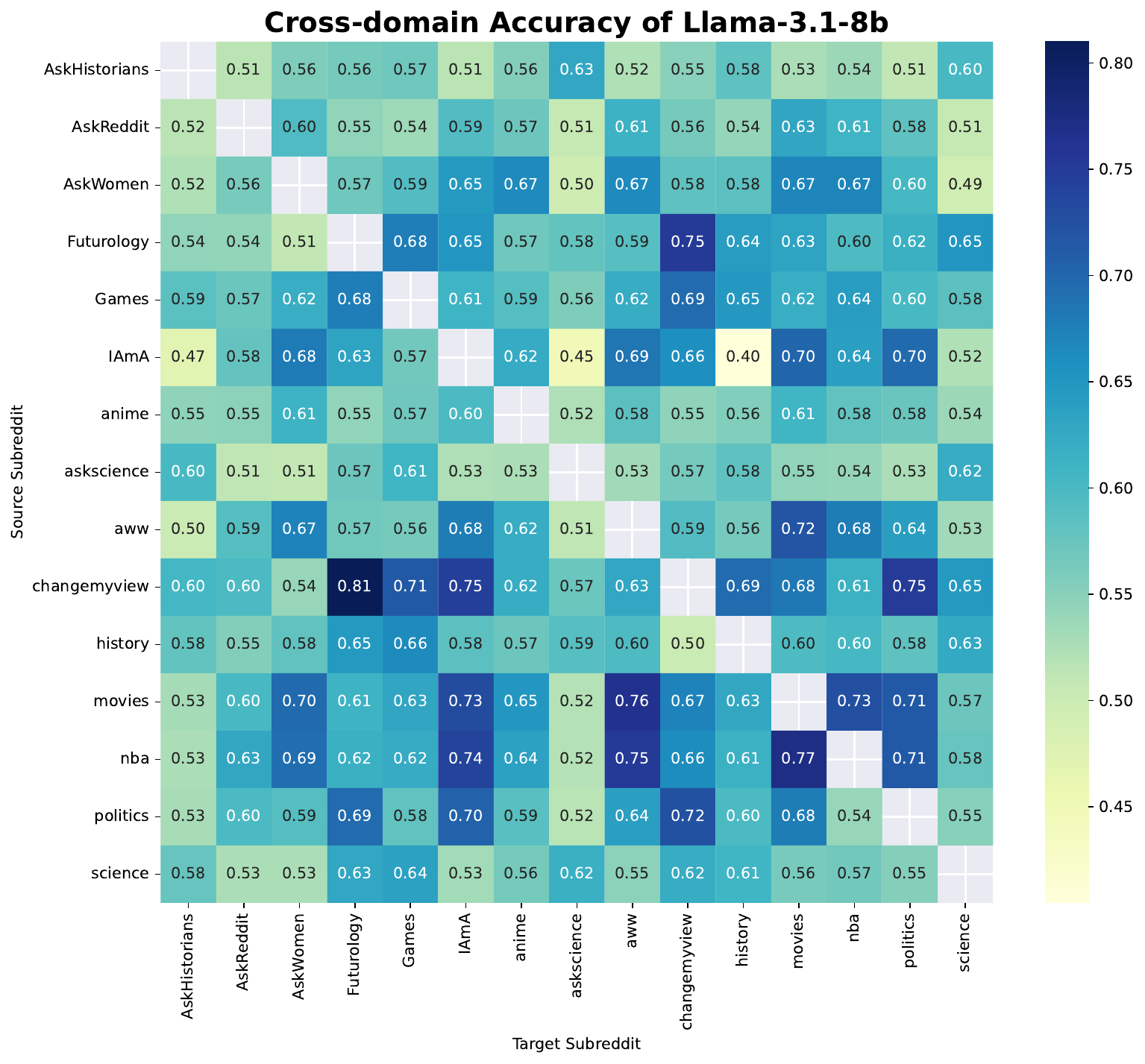} 
    \caption{\textbf{Cross-domain Moderation Performance for Llama-3.1-8b}.}
    \label{fig:cross_domain_llama}
\end{figure*}

\begin{figure*}[ht] 
    \centering
    \includegraphics[width=1\textwidth]{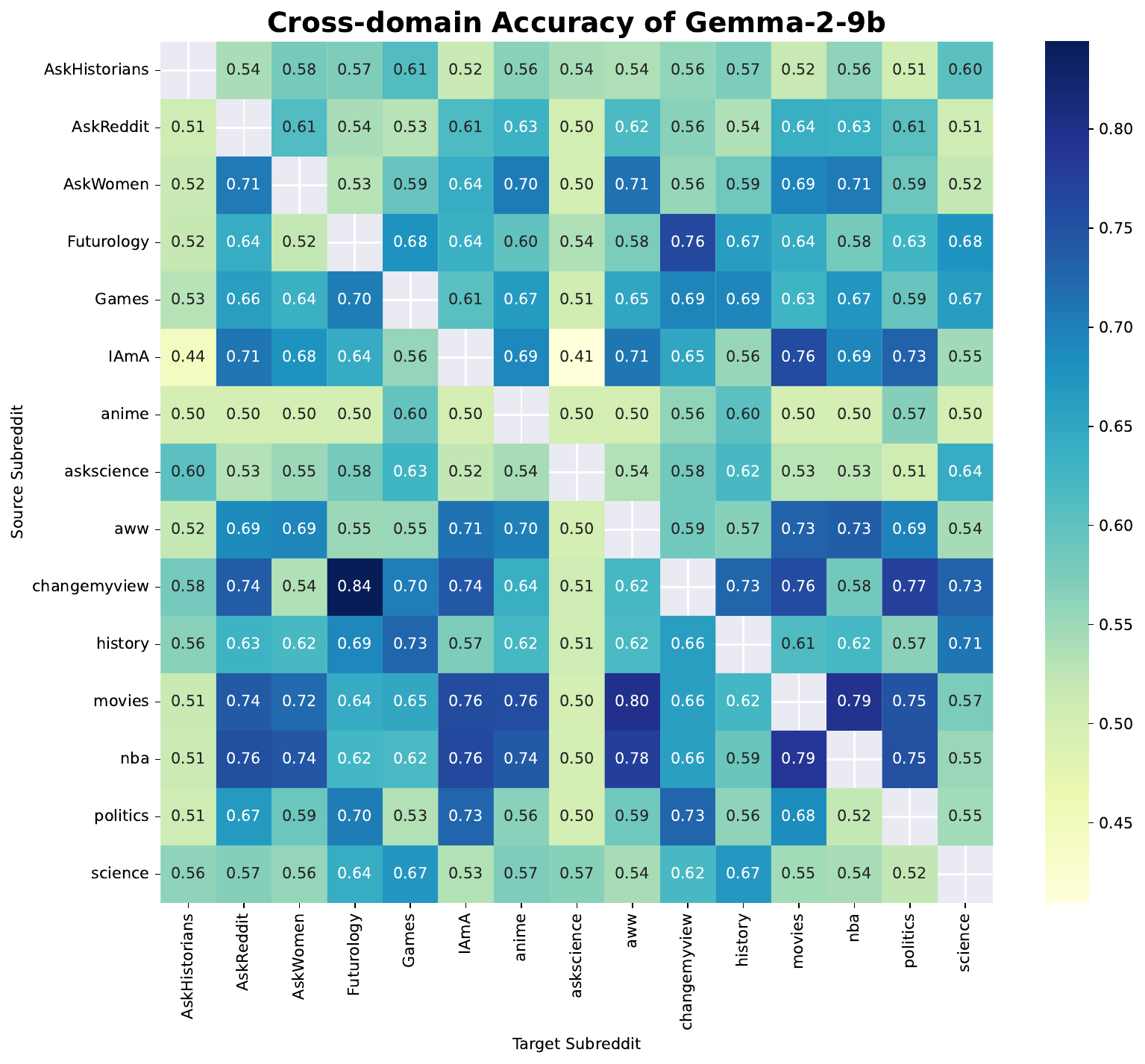} 
    \caption{\textbf{Cross-domain Moderation Performance for Gemma-2-9b}.}
    \label{fig:cross_domain_gemma}
\end{figure*}

\begin{figure*}[ht] 
    \centering
    \includegraphics[width=1\textwidth]{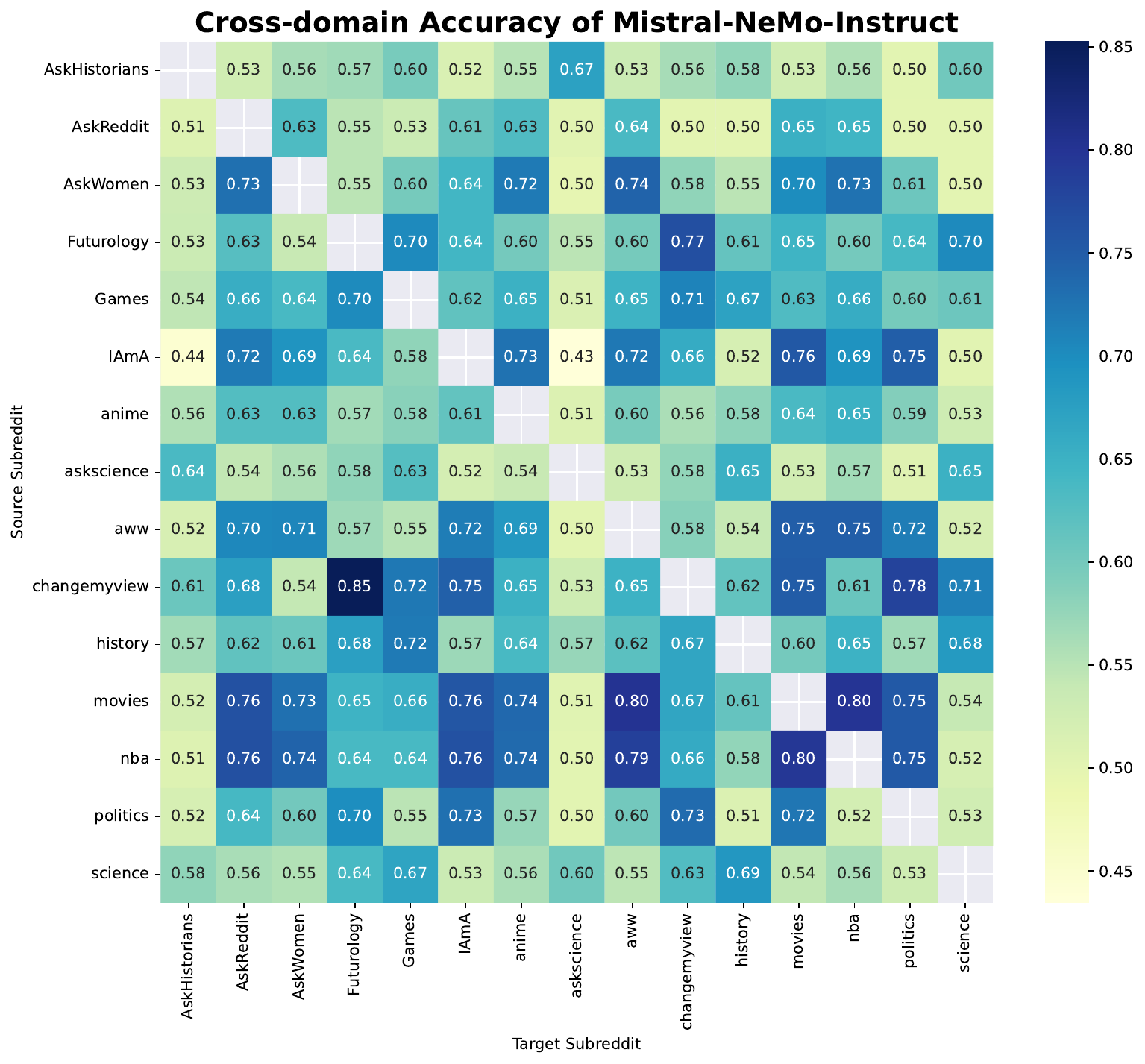} 
    \caption{\textbf{Cross-domain Moderation Performance for Mistral-NeMo-Instruct}.}
    \label{fig:cross_domain_mistral}
\end{figure*}

In this section we provide the model performances on cross-domain content moderation tasks across different subreddits. \autoref{fig:cross_domain_llama}, \autoref{fig:cross_domain_gemma}, and \autoref{fig:cross_domain_mistral} represent cross-domain accuracy of all subreddits for Llama-3.1-8b, Gemma-2-9b, and Mistral-NeMo-Instruct.

\section{Performance of Base SLMs on Content Moderation Task Without Fine-tuning}\label{app:base_model_performance}

In this section, we report the performance of base SLMs without fine-tuning on the in-domain content moderation tasks described in the main text. 

From \autoref{tab:base_slm_results} we observe that SLMs are ill-suited for content moderation prior to fine-tuning as compared to LLMs, which is expected given the difference in model sizes, capabilities, and lack of instruction-tuning. As a result, we decided to pursue fine-tuned SLMs for this work since they are more interesting to study. 

We observe that being an instruction-tuned model, Mistral-NeMo-Instruct performs the best and provides acceptable performance even prior to fine-tuning, but the performance of Llama-3.1-8b and Gemma-2-9b are near random chance.

\section{Detailed Cross-domain Correlation Test Results}\label{app:cross_domain_corr_test}

In this section, we provide a detailed set of results from the Pearson Correlation Coefficient t-test for non-correlation that we conducted to identify any patterns in cross-domain model performance. \autoref{tab:cross_domain_size_corr} shows test results for relative subreddit size, \autoref{tab:cross_domain_description_corr} shows the test results for embeddings of subreddit descriptions, and \autoref{tab:cross_domain_rules_corr} shows test results for embeddings of subreddit rules. Overall, we find only four instances of statistically significant correlation across all subreddits and models.

\begin{table*}[h]
\sffamily
\small
\centering
\caption{\textbf{Relative-Size Results of t-test for cross-domain setting.} Testing the null hypothesis of non-correlation between the relative sizes of the source and target subreddits with the cross-domain accuracy of SLMs. Values in the table represent the Pearson's Correlation Coefficient $r$ and statistically significant values with $p$-value < 0.05 are marked with \texttt{($\ast$)}. We see no statistically significant positive correlation.}
\label{tab:cross_domain_size_corr}
\begin{tabular}{l|ccc}
 \textbf{Subreddit} & \textbf{llama-3.1-8b} & \textbf{gemma-2-9b} & \textbf{mistral-nemo-instruct} \\
\hline
\textbf{r/AskHistorians} & -0.068 & -0.331 & -0.180 \\
\textbf{r/AskReddit} & 0.010 & -0.033 & 0.099 \\
\textbf{r/AskWomen} & 0.256 & 0.153 & 0.275 \\
\textbf{r/Futurology} & -0.340 & -0.421 & -0.388 \\
\textbf{r/Games} & -0.218 & -0.280 & -0.225 \\
\textbf{r/IAmA} & 0.033 & -0.106 & -0.036 \\
\textbf{r/anime} & 0.102 & -0.138 & 0.045 \\
\textbf{r/askscience} & -0.026 & -0.231 & -0.190 \\
\textbf{r/aww} & 0.073 & 0.063 & 0.033 \\
\textbf{r/changemyview} & -0.151 & -0.108 & -0.314 \\
\textbf{r/history} & -0.319 & -0.309 & -0.143 \\
\textbf{r/movies} & -0.029 & -0.003 & -0.080 \\
\textbf{r/nba} & 0.296 & 0.259 & 0.295 \\
\textbf{r/politics} & 0.062 & -0.033 & -0.101 \\
\textbf{r/science} & -0.309 & -0.287 & -0.291 \\
\bottomrule
\end{tabular}
\end{table*}

\begin{table*}[h]
\sffamily
\small
\centering
\caption{\textbf{Subreddit Description Embeddings Results of t-test for cross-domain setting.} Testing the null hypothesis of non-correlation between the between the cosine similarity matrix of pairwise subreddit rule embeddings between the source and target subreddits with the cross-domain accuracy of SLMs. Values in the table represent the Pearson's Correlation Coefficient $r$ and statistically significant values with $p$-value < 0.05 are marked with \texttt{($\ast$)}. We see only two instances of statistically significant positive correlation for Llama-3.1-8b on \textit{r/AskHistorians} and Gemma-2-9b on \textit{r/askscience}.}
\label{tab:cross_domain_description_corr}
\begin{tabular}{l|ccc}
\textbf{Subreddit} & \textbf{llama-3.1-8b} & \textbf{gemma-2-9b} & \textbf{mistral-nemo-instruct} \\
\hline
\textbf{r/AskHistorians} & 0.459 \texttt{($\ast$)} & 0.370 & 0.264 \\
\textbf{r/AskReddit} & 0.009 & -0.063 & -0.037 \\
\textbf{r/AskWomen} & -0.074 & 0.030 & -0.078 \\
\textbf{r/Futurology} & 0.099 & 0.152 & 0.116 \\
\textbf{r/Games} & 0.196 & 0.224 & 0.195 \\
\textbf{r/IAmA} & -0.004 & 0.045 & -0.020 \\
\textbf{r/anime} & 0.160 & 0.172 & 0.135 \\
\textbf{r/askscience} & 0.314 & 0.536 \texttt{($\ast$)} & 0.329 \\
\textbf{r/aww} & -0.070 & 0.138 & 0.042 \\
\textbf{r/changemyview} & 0.166 & 0.114 & 0.131 \\
\textbf{r/history} & 0.284 & 0.018 & -0.046 \\
\textbf{r/movies} & -0.168 & -0.086 & -0.139 \\
\textbf{r/nba} & -0.073 & -0.264 & -0.179 \\
\textbf{r/politics} & -0.155 & -0.202 & -0.371 \\
\textbf{r/science} & 0.377 & 0.273 & 0.367 \\
\bottomrule
\end{tabular}
\end{table*}

\begin{table*}[h]
\sffamily
\small
\centering
\caption{\textbf{Subreddit Rule Embeddings Results of t-test for cross-domain setting.} Testing the null hypothesis of non-correlation between the cosine similarity matrix of pairwise subreddit rule embeddings of the source and target subreddits with the cross-domain accuracy of SLMs. Values in the table represent the Pearson's Correlation Coefficient $r$ and statistically significant values with $p$-value < 0.05 are marked with \texttt{($\ast$)}. We see only two instances of statistically significant positive correlation for Llama-3.1-8b on \textit{r/nba} and Gemma-2-9b on \textit{r/anime}.}
\label{tab:cross_domain_rules_corr}
\begin{tabular}{l|ccc}
 \textbf{Subreddit} & \textbf{llama-3.1-8b} & \textbf{gemma-2-9b} & \textbf{mistral-nemo-instruct} \\
\hline
\textbf{r/AskHistorians} & 0.009 & 0.008 & -0.103 \\
\textbf{r/AskReddit} & 0.173 & -0.007 & 0.308 \\
\textbf{r/AskWomen} & 0.307 & 0.244 & 0.298 \\
\textbf{r/Futurology} & -0.116 & -0.058 & -0.057 \\
\textbf{r/Games} & 0.004 & -0.084 & -0.102 \\
\textbf{r/IAmA} & 0.025 & -0.021 & 0.065 \\
\textbf{r/anime} & 0.266 & 0.459 \texttt{($\ast$)} & 0.357 \\
\textbf{r/askscience} & -0.223 & 0.020 & -0.361 \\
\textbf{r/aww} & 0.257 & 0.201 & 0.229 \\
\textbf{r/changemyview} & 0.134 & -0.029 & -0.018 \\
\textbf{r/history} & 0.077 & -0.327 & -0.337 \\
\textbf{r/movies} & 0.344 & 0.262 & 0.322 \\
\textbf{r/nba} & 0.462 \texttt{($\ast$)} & 0.439 & 0.388 \\
\textbf{r/politics} & 0.416 & 0.454 & 0.330 \\
\textbf{r/science} & -0.054 & -0.007 & -0.010 \\
\bottomrule
\end{tabular}
\end{table*}

\end{document}